\documentclass[]{ceurart}
\usepackage{listings}
\begin{document}

\copyrightyear{2026}
\copyrightclause{Copyright for this paper by its authors.
  Use permitted under Creative Commons License Attribution 4.0
  International (CC BY 4.0).}

\conference{CLEF 2026 Working Notes, 21 -- 24 September 2026, Jena, Germany}

\title{DS@GT ARC at CheckThat! 2026: A Verdict-Anchored Pipeline for Generating Fact-Checking Articles}

\author[1]{Hoang Thanh Thanh Truong}[
    orcid=0009-0007-4130-3349,
    email=htruong47@gatech.edu,
]
\cormark[1]
\author[1]{Shuyu Tian}[
    orcid=0000-0001-6444-651X,
    email=stian40@gatech.edu,
]
\address[1]{Georgia Institute of Technology, North Ave NW, Atlanta, GA 30332}
\cortext[1]{Corresponding author.}

\begin{abstract}
Misinformation spreads faster than professional fact-checkers can manually verify it and write articles explaining their verdicts. This creates a need 
to automate the production of fact-checking articles. This paper details work done by the DS@GT ARC team for 
CLEF 2026 CheckThat! Lab Task 3 Fact-Checking Article Generation. We seek 
to generate a full fact-checking article from a claim, its original verdict, 
and a set of scraped source documents. Our team built a four-stage pipeline 
that cleans noisy evidence, retrieves and clusters relevant sources, generates 
the article by conditioning \texttt{Llama-3.1-8B-Instruct} on the original 
fact-checker verdict, and recovers inline citations through deterministic 
post-processing. We achieved a mean score of 0.428 and ranked 3rd out of 9 teams for the CLEF 2026 CheckThat! Lab Task 3, compared to a baseline mean score of 0.272.  Our code is available on GitHub at https://github.com/dsgt-arc/checkthat-2026-task3. 
\end{abstract}


\maketitle

\section{Introduction}

Misinformation travels faster than the truth. False news online has been shown to circulate more rapidly and reach a wider audience than verified information~\cite{vosoughi2018spread}. This rapid spread of misinformation online exceeds what professional fact-checkers, however diligent, can keep up with on their own. The resulting gap motivates the need for automated fact-checking systems.

Acknowledging the need to scale fact-checking beyond what human verifiers alone can sustain, the CheckThat! lab has organized a series of tasks targeting an automated fact-checking pipeline~\cite{clef-checkthat:2026-lncs}. The pipeline begins with detecting check-worthy claims, checks whether each claim has already been fact-checked, gathers relevant evidence, verifies the claim against that evidence, and generates a brief explanation of the verdict. Sahnan et al.~\cite{sahnan2025qraft} build on this pipeline structure and propose adding one more step: generating the full fact-checking article that professional fact-checkers publish for the public. Accordingly, the CheckThat! lab extends its pipeline in the 2026 cycle to include this final stage as a task.

This stage corresponds to CheckThat! 2026 Task 3 ~\cite{clef-checkthat:2026:task3}, formalized as Fact-Checking Article Generation. Given a claim, its original verdict, and a set of source documents scraped from online sources, participants must produce a fact-checking article that entails the gold reference written by professional fact-checkers (entailment), cites the provided sources both accurately (citation precision) and comprehensively (citation recall), and covers all of the supplied evidence (evidence coverage). Systems are evaluated along these four metrics, whose mean determines the final leaderboard ranking.

In this paper, we describe the DS@GT ARC at CheckThat! submission to CheckThat!~2026 Task~3. The system is a four-stage pipeline consisting of evidence cleaning, evidence retrieval and clustering, article generation, and citation post-processing. Rather than asking the model to write an open-ended article from evidence alone, we built a verdict-anchored generation pipeline that conditions \texttt{Llama-3.1-8B-Instruct} on the original fact-checker verdict so the model writes toward a known conclusion. This submission ranked third out of nine participating teams on the official leaderboard with a mean score of 0.428, well above the baseline of 0.272~\cite{clef-checkthat:2026:task3}.

\section{Related Work}

Automated fact-checking has historically focused on verdict prediction, with later work extending to short justification generation~\cite{guo-etal-2022-survey}. However, Guo et al. observe that justification generation has received less attention than the verdict-prediction stages of the pipeline, despite being essential to communicating verdicts to readers. 

In addition to justification generation, the field has turned to another task: the generation of full fact-checking articles. Sahnan et al.~\cite{sahnan2025qraft} recently proposed this as a distinct task, taking the pipeline one step further toward the long-form output that professional fact-checkers could publish. 

This submission contributes to this emerging line of work, building on the claim-verification benchmarks and justification-generation systems that preceded it. Each task builds on the one before it. Claim-verification benchmarks supply the data and models, justification generation adds short explanations of the verdict, and article generation produces the full long-form output that fact-checkers could publish.

\subsection{Claim-verification Benchmarks}

Automated fact-checking has traditionally been framed as a claim-verification problem. Given a claim and a body of evidence, the system assigns a degree of truthfulness. Earlier work by Vlachos and Riedel~\cite{vlachos-riedel-2014-fact} framed fact-checking as ordinal classification on a multi-point scale, while FEVER~\cite{thorne-etal-2018-fever} introduced the categorical three-way scheme (supported, refuted, or not enough info) that has since become standard.

Building on this three-way framework, subsequent benchmarks expanded the claim-verification task along other dimensions, particularly in the kinds of claims and evidence they used. While FEVER used crowd-sourced claims from Wikipedia sentences, MultiFC~\cite{augenstein-etal-2019-multifc} drew claims from real-world fact-checking organizations. This shift moved the task closer to the kinds of claims fact-checkers actually verify in practice.

Extending this real-world setting, WatClaimCheck~\cite{khan-etal-2022-watclaimcheck} provides premise articles, which are the actual source documents that professional fact-checkers consulted while writing their review. The CheckThat! 2026 Task 3 training data are drawn from WatClaimCheck, which makes the premise-article setting a natural fit for article-generation work that needs to ground its output in the same source material the professional fact-checkers used.

\subsection{Justification Generation}

A growing line of work has moved beyond verdict prediction toward producing natural-language justifications of the verdict. For instance, Atanasova et al.~\cite{atanasova-etal-2020-generating-fact} approached justification generation as a summarization problem. Given a full fact-check article as input, their model selects a few sentences from it to form a short justification of the verdict. 

However, this setup assumes the fact-check article already exists. This is unrealistic, since fact-check articles are only written after a claim has been verified. JustiLM~\cite{zeng-gao-2024-justilm} relaxed this assumption. They generate justifications from retrieved evidence documents instead of from the fact-check article. The fact-check article is used only as an auxiliary training signal.

In the same study, Zeng and Gao also introduced the ExClaim benchmark, which they constructed by extracting the conclusive paragraphs of WatClaimCheck's fact-check articles.  ExClaim later became the test split for CheckThat! 2026 Task 3.

The justification-generation studies differ from the developments in this competition in scope. Justifications are short rationales, roughly a paragraph long, summarizing why the verdict was reached. By contrast, Task~3 requires producing a full fact-checking article in the style of articles published by professional fact-checkers. Justification generation shares with article generation the challenge of producing text faithful to the supporting evidence. However, it stops well short of the structural and narrative demands of a complete article.

\subsection{Full Fact-Checking Article Generation}

Unlike claim verification and justification generation, full fact-checking article generation has only recently emerged as a distinct task. Sahnan et al.~\cite{sahnan2025qraft} argue that the automatic fact-checking pipeline, which ends with brief explanation generation, omits the actual deliverable that professional fact-checkers produce. This deliverable is a long-form article suitable for public dissemination. They propose QRAFT, a multi-agent framework with three agents. A Planner agent extracts evidence nuggets and proposes an outline. A Writer agent drafts the article. An Editor agent iteratively refines the draft through conversational question-asking.

QRAFT, built on GPT-4o and GPT-4o-mini, outperforms several strong baselines on automatic metrics, including JustiLM~\cite{zeng-gao-2024-justilm} and a zero-shot GPT-4o-mini baseline. However, when professional fact-checkers compared QRAFT's articles to human-written ones, they still preferred the latter. 

The CheckThat! 2026 Task 3~\cite{clef-checkthat:2026:task3} builds on this task definition. Unlike QRAFT, which uses human evaluation and LLM-as-a-judge with a reader-focused rubric, Task 3 adopts an automatic evaluation protocol. It measures entailment with the gold reference, citation precision and recall, and evidence coverage. This enables comparison across many submitted systems but shifts the focus from reader-perceived article quality toward faithfulness to the provided sources.

This submission contrasts with QRAFT in two main ways, each motivated by a different notion of practicality. First, we use a frozen open-source generator (\texttt{Llama-3.1-8B-Instruct}) rather than a proprietary API model. This reduces inference cost by an order of magnitude and makes the system reproducible without API access, in contrast to QRAFT's reliance on GPT-4o and GPT-4o-mini. Second, instead of QRAFT's iterative multi-agent loop, which repeatedly invokes a Planner, Writer, and Editor agent to refine each article, we use a single-pass pipeline built around two lightweight, deterministic components: a verdict-anchored prompt and a rule-based citation-recovery post-processor. Where QRAFT scales complexity by adding more agents and more model calls, our pipeline scales by adding structure around a single generation pass, cleaning and organizing evidence before generation and correcting citation formatting after it. 

The novelty of this approach lies not in a new architecture or training method, but in testing the extent to which careful, deterministic engineering around a small open-source model can close the gap to methods that rely on repeated calls to larger proprietary models. This makes the pipeline considerably cheaper to run and easier for other teams to reproduce. As our ablation results show, however, this engineering primarily improves citation quality and evidence coverage rather than the model's underlying reasoning, which suggests that more adaptive mechanisms, such as an agentic loop, may still be necessary to close the remaining gap in entailment.

\section{Methodology}

The task is to take a claim, its original verdict, and a set of source documents as input, and produce a fact-checking article with inline citations as output. The pipeline consists of four stages: evidence cleaning, evidence retrieval and clustering, article generation, and citation post-processing. We first describe the official baseline against which we compare, then present our exploratory data analysis, and finally detail each stage of our pipeline.

\subsection{Baseline Methods}

The official CheckThat!~2026 Task~3 baseline~\cite{clef-checkthat:2026:task3} prompts \texttt{GPT-5-mini} in a zero-shot manner to generate fact-checking articles. For each instance, the model receives the claim, the claimant, the original verdict, and the evidence excerpts truncated to 4000 characters per source. It is instructed to produce a complete article with a headline, detailed analysis, and a final verdict, with inline citations in the format \texttt{(source: <url>)}. The baseline applies no preprocessing to the source documents and no post-processing to the generated article. We use this baseline as the primary point of comparison for the submission. 

\subsection{Exploratory Data Analysis}

The CheckThat!~2026 Task~3 dataset~\cite{clef-checkthat:2026:task3} provides a training split drawn from WatClaimCheck and a test split drawn from ExClaim. The training split contains 3{,}372 instances. The organizers additionally release a 200-instance subset of the training split with corresponding baseline-generated articles produced by \texttt{GPT-5-mini}. We use this 200-instance subset throughout development to compare the pipeline against the baseline on identical inputs. Final leaderboard results are reported on the held-out test split, where both the submission and the official baseline are scored.

Each instance includes a claim, an original verdict assigned by professional fact-checkers, and a set of premise articles cited in the original fact-check. The training split uses JSON-formatted evidence files. The test split contains JSON files alongside plain-text evidence files from ExClaim.

During the development phase, we characterize the data using the 200-instance development subset, which exactly matches the inputs the baseline was scored on. Claims average 17 words in length. The verdicts span 23 distinct fine-grained labels used by the source fact-checking organizations, which the dataset coarsens into three classes: false (46.5\%), true (44.0\%), and mixture (9.5\%). The number of premise articles per claim varies considerably, ranging from 2 to 94 with a median of 6. Across the subset, premise articles total 1{,}513 source documents. 

Article length varies even more widely. The median article contains 867 words, but lengths range from 1 to over 148{,}000 words. This high-variance, large-context setting motivates both the evidence cleaning step, which filters empty and boilerplate sources, and the retrieval-and-clustering step, which selects and organizes evidence to fit within the model's context window. 

\subsection{Our Pipeline}

\begin{figure}[t]
  \centering
  \includegraphics[width=0.3\linewidth]{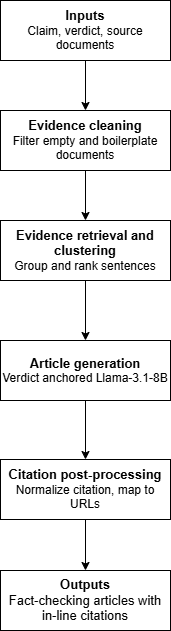}
  \caption{Overview of the DS@GT ARC pipeline: evidence cleaning, retrieval and clustering, verdict-anchored article generation, and citation post-processing.}
  \label{fig:pipeline}
\end{figure}

Our pipeline consists of four stages: evidence cleaning, evidence retrieval and clustering, article generation, and citation post-processing. Figure~\ref{fig:pipeline} illustrates this pipeline, and each stage is described in detail below.

\subsubsection{Evidence Cleaning}

The premise articles vary widely in quality. Some documents contain clean article text, while others contain only website boilerplate such as navigation menus, cookie banners, and paywall messages. In addition, some documents are effectively empty, containing little usable content. We address this variation by classifying each source document into one of three categories before constructing the prompt:
\begin{itemize}
\item \textbf{Empty}: documents whose raw word count is zero or smaller than the claim's word count, on the assumption that a document shorter than the claim cannot contain meaningful evidence.
\item \textbf{Boilerplate}: non-empty documents with low semantic overlap with the claim. We measure this overlap as the cosine similarity between the claim and the document, computed with \texttt{all-MiniLM-L6-v2}~\cite{reimers-gurevych-2019-sentence} sentence embeddings, and mark documents below a threshold of 0.25 as boilerplate.
\item \textbf{Relevant}: the remaining documents, which are non-empty and exceed the similarity threshold.
\end{itemize}

For long premise articles, we do not explicitly chunk or truncate the text before embedding. The full cleaned document is passed to \texttt{all-MiniLM-L6-v2}, which applies its own fixed input-length limit (256 tokens) internally. As a result, similarity for very long documents is computed primarily from their opening content, since text beyond the model's input limit is not seen during embedding. We did not evaluate alternative aggregation strategies, and note this as a limitation of the current pipeline.

Only relevant documents are forwarded to the retrieval stage. Their text is cleaned by stripping HTML markup, transliterating to ASCII, removing non-printable characters, and deduplicating sentences. Empty and boilerplate documents are excluded from the LLM prompt but tracked separately and reintroduced after generation as a deterministic appendix sentence of the form:
\begin{quote}
\textit{The following sources were consulted but could not be accessed (source: A; B; C).}
\end{quote}
This guarantees that every provided source filename appears in the article's citation set without polluting the LLM's working context with content-free material. This design choice is explicitly aimed at improving citation recall and evidence coverage as measured by the automatic scorer. It does not reflect genuine engagement with those sources' content, since the excluded documents contain little or no usable evidence in the first place.

\subsubsection{Evidence Retrieval and Clustering}

For each claim, we take the relevant documents identified in the previous stage and organize their evidence in two steps. We first cluster documents that discuss the same factual point. We then select the most claim-relevant sentences within each cluster, using the same encoder as in the evidence cleaning stage.

We cluster the relevant documents by semantic similarity. We embed each document's cleaned text, compute pairwise cosine distances, and apply agglomerative clustering with average linkage and a cosine-distance threshold of 0.5. Each resulting cluster groups documents that cover the same factual point. Clustering serves only as an internal selection mechanism. It determines which sentences are kept, but the model itself never sees the cluster structure. After sentence selection, the cluster groupings are discarded, and the chosen sentences are presented to the model as a single flat list, organized by source.

Within each cluster, we select sentences by relevance to the claim. We split each document into sentences, discard fragments shorter than five words, and score each sentence by its cosine similarity to the claim. To ensure no source is silently dropped, we first reserve the single highest-scoring sentence from every source in the cluster. We then fill the remaining budget, up to ten sentences per cluster, from the ranked pool of remaining sentences.

We set several thresholds in this stage heuristically, rather than through a systematic hyperparameter search: a boilerplate similarity cutoff of 0.25, a clustering distance threshold of 0.5, a maximum of ten sentences per cluster, and a minimum sentence length of five words. We based these choices on manual inspection of the development subset, and leave a more thorough sensitivity analysis to future work.

\subsubsection{Article Generation}

We generate articles with \texttt{Llama-3.1-8B-Instruct}~\cite{grattafiori2024llama3herdmodels}, 
loaded via the HuggingFace Transformers library. For each claim, 
the prompt instructs the model to act as a fact-checker and to write a complete 
article using only the provided evidence. The prompt presents the claim, its 
metadata, and the clustered evidence in a fixed template:

\begin{verbatim}
[Instruction explaining the task]
Claim: {claim}
Claimant: {claimant}
Original rating: {original_rating}
Evidence articles:
[{filename_1}]: {selected sentences}
[{filename_2}]: {selected sentences}
...
\end{verbatim}

\paragraph{Inference settings.}
Articles are generated with \texttt{meta-llama/Llama-3.1-8B-Instruct}, loaded in \texttt{bfloat16} precision via the HuggingFace Transformers \texttt{pipeline} API. We generate up to 1024 new tokens per article. Generation uses sampling with a temperature of 0.2 rather than greedy decoding. Top-\emph{p} and top-\emph{k} are left at their HuggingFace defaults rather than explicitly tuned, and no repetition penalty is applied. We do not impose an explicit maximum input length. Prompts are passed to the tokenizer without truncation, relying on the model's native context window. Articles are generated one claim at a time rather than in batches. We did not perform a systematic search over these settings. The low sampling temperature was chosen to keep generation close to deterministic while still allowing some variation across runs.

\paragraph{Verdict anchoring.}
Rather than asking the model to derive a verdict from the evidence alone, we provide the original fact-checker rating as an input field. This conditions the model to write toward a known conclusion. The rating acts as a target verdict for the generated narrative. We expect this to improve the article's logical consistency with the gold reference, and in turn its entailment score. We refer to this as \emph{verdict anchoring}.

\paragraph{Citation format enforcement.}
The prompt specifies the required citation format. The model is instructed to place an inline citation of the form \texttt{(source: <filename>)} at the end of any sentence that draws on evidence. When several documents support the same point, it cites them together as \texttt{(source: <fn1>; <fn2>; ...)}. It is explicitly told not to invent sources and to cite only the provided filenames. The official scorer only recognizes citations in this format and discards any others during scoring. Despite the explicit instruction, the model still emits malformed citations that the scorer cannot recognize. These are corrected in the post-processing stage described next.

\paragraph{Inaccessible sources.}
The empty and boilerplate sources excluded during cleaning are reintroduced after generation as a deterministic appendix sentence. Thus, every supplied source filename appears in the article's citation set.

\subsubsection{Citation Post-processing}

The generation stage produces citations that reference source \emph{filenames}, 
but the Task~3 scorer expects citations to reference source \emph{URLs}. The 
model also occasionally deviates from the canonical \texttt{(source: <filename>)} 
form. We correct these issues with a deterministic, rule-based post-processing 
pass that applies the following steps.

\begin{enumerate}
    \item \textbf{Malformed wrappers.} Citations with mismatched or doubled 
    delimiters, such as \texttt{((source: X))}, \texttt{[source: X)}, and 
    \texttt{(source: X]}, are normalized to \texttt{(source: X)}.
    \item \textbf{Nested citations.} Doubly nested forms such as 
    \texttt{(source: (source: X))} are flattened to \texttt{(source: X)}.
    \item \textbf{Bracket variants.} Square-bracket references 
    \texttt{[filename.json]} are rewritten as \texttt{(source: filename.json)}.
    \item \textbf{Missing citations.} If a source filename appears in the body 
    of the article but is never cited in a \texttt{(source: \ldots)} marker, we 
    append \texttt{(source: filename)} as a standalone sentence at the end of 
    the article. This is conservative. It does not attribute any specific claim 
    to the source, but it ensures the filename enters the citation set so that 
    it counts toward recall and coverage.
    \item \textbf{Filename-to-URL substitution.} Finally, each filename inside a 
    \texttt{(source: \ldots)} citation is replaced with its corresponding source 
    URL, looked up from the dataset's premise-article metadata. Multi-source 
    citations are split on semicolons and commas before substitution.
\end{enumerate}

This pass recovers citations that the scorer would otherwise discard, directly 
improving citation precision, citation recall, and evidence coverage. Like the inaccessible-sources appendix, this post-processing pass is explicitly designed to improve the citation-based metrics as measured by the automatic scorer. It does not verify that a citation is used correctly or that the cited evidence genuinely supports the sentence it is attached to. It only ensures that the citation is present and correctly formatted so that it is counted by the scorer.

\section{Results}

\subsection{Main Result}

Table~\ref{tab:main} reports our submission on the held-out test split of 1{,}158 
instances, alongside the official baseline. Our system 
achieves a mean score of 0.428, ranking third out of nine participating teams, 
well above the baseline mean of 0.272.

\begin{table}[t]
\centering
\caption{Results on the Task~3 test split. Mean is the average of the four metrics.}
\label{tab:main}
\begin{tabular}{lccccc}
\toprule
System & Entail. & Cite Prec. & Cite Rec. & Ev. Cov. & Mean \\
\midrule
Baseline             & 0.298 & 0.223 & 0.240 & 0.329 & 0.272 \\
\textbf{DS@GT (3rd)} & 0.295 & 0.240 & 0.276 & \textbf{0.902} & \textbf{0.428} \\
\bottomrule
\end{tabular}
\end{table}

Our improvement over the baseline is concentrated in a single metric: evidence coverage. We score 0.902 against the baseline's 0.329. On the other three metrics, the two systems are comparable. Our entailment (0.295) is marginally below the baseline's (0.298). Our improvement in mean score is therefore almost entirely a coverage gain. 

This is notable because the baseline uses a far larger proprietary model (\texttt{GPT-5-mini}) than our 8B open-source generator. The difference is not that our model writes better prose, but that our pipeline surfaces and integrates far more of the supplied evidence, through evidence cleaning, the per-source sentence guarantee, and the inaccessible-sources appendix. The baseline, prompting the model directly with truncated evidence, leaves most sources uncovered. Entailment, by contrast, stays low for both systems. These findings suggest that faithfully reproducing the gold reference's reasoning is the hard part of the task and is not addressed by evidence-handling alone.

\subsection{Ablation Study}

To understand which design choices drive performance, we evaluate four configurations on the 200-instance development subset, on which the official baseline articles are also provided. All configurations share the same evidence cleaning and citation post-processing. The base configuration uses a plain generation prompt, while the remaining three use the verdict-anchored prompt. The latter three also differ in their retrieval encoder: the default \texttt{all-MiniLM-L6-v2}, a cross-encoder reranker, and BGE embeddings. Table~\ref{tab:ablation} reports the results.

\begin{table}[t]
\centering
\caption{Ablation on the 200-instance development subset. Mean is the average of the four metrics.}
\label{tab:ablation}
\begin{tabular}{lccccc}
\toprule
Configuration & Entail. & Cite Prec. & Cite Rec. & Ev. Cov. & Mean \\
\midrule
Base (plain prompt)        & 0.298 & 0.051 & 0.138 & 0.765 & 0.313 \\
\;+ verdict anchoring      & 0.311 & 0.210 & 0.304 & 0.894 & 0.430 \\
\;\;+ cross-encoder rerank & 0.309 & 0.213 & 0.321 & 0.891 & \textbf{0.433} \\
\;\;+ BGE embeddings       & 0.309 & 0.213 & 0.294 & 0.889 & 0.426 \\
\bottomrule
\end{tabular}
\end{table}

\paragraph{Verdict anchoring is the dominant lever.}
Replacing the base prompt with the verdict-anchored prompt raises the mean score from 0.313 to 0.430, an increase of 0.117. The improvement is concentrated in the citation metrics. Citation precision rises from 0.051 to 0.210 and recall from 0.138 to 0.304. Evidence coverage rises from 0.765 to 0.894. The effect on entailment is small, from 0.298 to 0.311. This indicates that conditioning generation on the original verdict primarily helps the model produce well-formed, well-grounded citations. It does not substantially improve the article's logical alignment with the gold reference.

\paragraph{The retrieval encoder makes little difference.}
On top of verdict anchoring, the choice of retrieval encoder has only a marginal effect. Cross-encoder reranking gives a small gain, from 0.430 to 0.433, driven by slightly higher citation recall. BGE embeddings slightly reduce the mean score, from 0.430 to 0.426. We therefore retain the default \texttt{all-MiniLM-L6-v2} encoder for our submission, as the more expensive alternatives do not yield a consistent improvement.

\section{Discussion}

The ablation results in Table~\ref{tab:ablation} show that verdict anchoring is the dominant lever in this pipeline. Moving from the base (plain prompt) configuration to the verdict-anchored prompt raises the four-metric mean from 0.313 to 0.430. Nearly all of that gain is concentrated in the citation metrics rather than entailment. This finding suggests that, without the original verdict, the model may need to implicitly infer a conclusion from the evidence before it can write toward it. This added burden could compete with the model's ability to track and format citations correctly. 

Supplying the verdict upfront may remove some of that inference step, potentially freeing the model to focus more of its capacity on grounding sentences in the correct sources rather than deciding what to conclude. This is consistent with the small movement in entailment (0.298 to 0.311) alongside the much larger movement in citation precision (0.051 to 0.210) and recall (0.138 to 0.304): verdict anchoring changes how well the model cites evidence far more than it changes how well the generated narrative matches the gold reference's reasoning.

This distinction matters for interpreting our main result in Table~\ref{tab:main}. On the held-out test split, our entailment score (0.295) is comparable to the baseline's (0.298) despite our much larger mean-score lead (0.428 vs. 0.272). Most of the improvement appears to come from evidence coverage (0.902 vs. 0.329) with smaller gains in citation precision and recall. In other words, our pipeline is substantially better at citing and covering the supplied evidence, but it is not demonstrably better at producing an article whose reasoning matches what professional fact-checkers concluded. Verdict anchoring narrows the gap on citation quality. It does not appear to be a mechanism for improving factual reasoning itself.

The citation post-processing stage (Section~3.3.4) plays a complementary but distinct role. Because it is a deterministic, rule-based correction applied after generation, it recovers citations that the scorer would otherwise discard due to malformed brackets, nested wrappers, or missing filename-to-URL substitution. This kind of correction is, by construction, unable to affect entailment, since it only touches citation formatting rather than the article's underlying content or claims. Its contribution is therefore best understood as protecting the citation-recall and evidence-coverage gains already produced by verdict anchoring and clustering, rather than as an independent driver of quality.

Taken together, these results suggest that our engineering choices, evidence cleaning, per-source sentence guarantees, verdict anchoring, and citation post-processing are effective at ensuring the model uses and correctly cites the evidence it is given. However, they do not address the harder problem of aligning the model's reasoning with the gold reference. Closing that gap likely requires a different approach than further refinements to retrieval or citation handling. Instead, progress may come from an explicit agentic mechanism, such as an AI agent that evaluates and revises the generated narrative against the evidence.

\section{Future Work}

Our ablation results show that verdict anchoring, which conditions the article on the correct verdict, helped the model cite evidence more accurately than changes to the retrieval encoder did. However, this improvement did not extend to entailment, which remained comparable to the baseline. Closing this gap likely requires directions beyond retrieval and prompting alone.

One direction is to scale up the generator itself. Although we did not test different generator sizes in this work, our results suggest that the 8-billion parameter model may struggle to balance synthesizing new information with following the required citation format. An immediate next step is to try AWQ-quantized versions of \texttt{Llama-3.1-70B-Instruct} and \texttt{Mistral-Large-Instruct-2407}, which could improve performance without a proportional increase in inference cost.

A second direction is an explicit agentic mechanism to help close the gap between citation quality and entailment. Future iterations of the pipeline could use an agentic loop with several LLM roles. For example, one model could cross-examine the retrieved evidence, another could plan the article's argument, another could write the draft, and another could edit it. These roles could iteratively revise both the content and formatting of the article before it is finalized. This would let the system check and correct its own reasoning against the evidence, rather than relying on a single generation pass.

\section{Conclusions}

This work presents a four-stage pipeline for CheckThat! 2026 Task 3, combining evidence cleaning, retrieval and clustering, verdict-anchored generation, and citation post-processing. The pipeline achieved a four-metric mean score of 0.428 on the held-out test split, ranking third out of nine teams and outperforming the official baseline score of 0.272.

This result shows that targeted preprocessing, verdict anchoring, and rule-based citation recovery are effective at improving citation precision, citation recall, and evidence coverage, even when using a small open-source generator. However, our ablation results also show that these techniques had little effect on entailment, which remained comparable to the baseline. This suggests that faithfully reproducing the gold reference's reasoning remains an open challenge that our pipeline's engineering choices do not address, and one that we identify as a priority for future work.

\begin{acknowledgments}
We thank the Data Science at Georgia Tech (DS@GT) CLEF competition group for their support.
This research was supported in part through research cyberinfrastructure resources and services provided by the Partnership for an Advanced Computing Environment (PACE) at the Georgia Institute of Technology, Atlanta, Georgia, USA \cite{PACE}.
\end{acknowledgments}

\section*{Declaration on Generative AI}
 
 During the preparation of this work, the author(s) used Claude in order to: grammar and spelling check, content brainstorming, and writing data-analysis and evaluation scripts. After using these tool(s)/service(s), the author(s) reviewed and edited the content as needed and take(s) full responsibility for the publication's content.

\bibliography{main}
\end{document}